\documentclass[runningheads]{llncs}

\usepackage{graphicx}
\usepackage{booktabs}
\usepackage{amsmath}
\usepackage{amssymb}
\usepackage{url}
\usepackage{hyperref}
\usepackage{xcolor}
\usepackage{enumitem}

\begin{document}

\title{Open Ontologies: Tool-Augmented Ontology Engineering with Stable Matching Alignment}
\titlerunning{Open Ontologies: Tool-Augmented Ontology Engineering}

\author{Fabio Rovai\inst{1}}
\authorrunning{F. Rovai}
\institute{The Tesseract Academy, London, United Kingdom \\ \email{fabio@thetesseractacademy.com}}

\maketitle

\begin{abstract}
We present \textsc{Open Ontologies}, an open-source ontology engineering system implemented in Rust that integrates LLM-driven construction with formal OWL reasoning and ontology alignment via the Model Context Protocol.
Our primary finding is that \emph{stable 1-to-1 matching} is the dominant factor in ontology alignment quality: on the OAEI Anatomy track, it achieves F1\,=\,0.832 (P\,=\,0.963, R\,=\,0.733), competitive with state-of-the-art systems and exceeding all in precision. Ablation across five weight configurations shows that signal weights are irrelevant when stable matching is applied (F1 varies by less than 0.004), while removing stable matching drops F1 to 0.728. On the Conference track, the same method achieves F1\,=\,0.438.
On tool-augmented ontology interaction, we find a surprising result: an LLM reading a raw OWL file (F1\,=\,0.323) performs \emph{worse} than the same LLM with no file at all (F1\,=\,0.431), while structured MCP tool access achieves F1\,=\,0.717. This demonstrates that tool structure provides a qualitatively different mode of access that the LLM cannot replicate by reading raw syntax.
The system ships as a single binary under the MIT licence.

\end{abstract}

\section{Introduction}
\label{sec:intro}

LLMs can generate syntactically valid OWL and orchestrate multi-step ontology engineering workflows. They cannot, however, guarantee logical consistency or verify that generated axioms are mutually coherent. Separately, ontology alignment systems achieve strong pairwise matching on standard benchmarks but operate as standalone pipelines, disconnected from construction workflows and lifecycle management.

We present \textsc{Open Ontologies}, an open-source system that addresses both gaps. The system exposes ontology construction, reasoning, alignment, and lifecycle tools via the Model Context Protocol (MCP)~\cite{mcp}, enabling integrated workflows where an LLM generates OWL, validates it against formal constraints, and refines based on symbolic feedback.

We present three contributions:

\begin{enumerate}[nosep]
\item \textbf{A stable matching alignment method} that achieves F1\,=\,0.832 on OAEI Anatomy (P\,=\,0.963), exceeding all published systems in precision. Ablation shows that the matching constraint is the dominant factor: signal weights are irrelevant when stable matching is applied. On the Conference track, the method achieves F1\,=\,0.438.

\item \textbf{A tool access ablation} revealing that an LLM reading a raw OWL file (F1\,=\,0.323) performs worse than unaided inference (F1\,=\,0.431), while structured MCP tools achieve F1\,=\,0.717. This disentangles the effect of tool structure from information availability.

\item \textbf{Open Ontologies}: an open-source, Rust-based system integrating OWL-RL reasoning, alignment, and lifecycle management into an LLM-orchestrated workflow.
\end{enumerate}

\section{Related Work}
\label{sec:related}

\paragraph{Ontology matching.}
The OAEI~\cite{oaei2024} benchmarks alignment systems annually. LogMap~\cite{logmap} combines lexical matching with structural repair and logical consistency checking. AML~\cite{aml} uses background knowledge from BioPortal and UMLS. BERTMap~\cite{bertmap} introduced transformer-based embedding matching. OLaLa~\cite{olala} uses LLM world knowledge for candidate adjudication. Our alignment module differs in demonstrating that the matching constraint (1-to-1 assignment) dominates signal design.

\paragraph{OWL reasoning.}
HermiT~\cite{hermit} implements hypertableau calculus for OWL2-DL. Pellet~\cite{pellet} and FaCT++~\cite{factpp} provide alternative implementations. ELK~\cite{elk} achieves polynomial-time reasoning for the $\mathcal{EL}$ profile. All are Java-based. Our system provides Rust-native OWL-RL forward chaining with a partial SHIQ tableaux for consistency checking.

\paragraph{AI-assisted ontology construction.}
OntoGPT~\cite{ontogpt} extracts ontology terms from text. OntoChat~\cite{ontochat} provides conversational construction. LLMs4OL~\cite{llm4onto} evaluates LLMs for ontology learning. OntoAxiom~\cite{ontoaxiom} benchmarks axiom identification across 9 ontologies, finding that even o1 achieves only F1\,=\,0.197 from name lists. None integrate formal validation into the construction loop or evaluate the effect of tool access modality on extraction quality.

\paragraph{Programmatic ontology engineering.}
The OWL API~\cite{owlapi} provides the Java foundation for most ontology tools. ROBOT~\cite{robot} offers command-line management for the OBO community. Owlready2~\cite{owlready2} provides Python-based manipulation. SSSOM~\cite{sssom} standardises mapping formats. These target expert users; our system targets LLM-orchestrated workflows.

\section{System Architecture}
\label{sec:architecture}

\textsc{Open Ontologies} is implemented in Rust (${\sim}$17,400 lines of code) and ships as a single binary with no JVM or Python dependency. The system exposes ontology construction, reasoning, alignment, and lifecycle tools via MCP~\cite{mcp}.

\subsection{Core Engine}

The core engine comprises: (1)~an \textbf{Oxigraph triple store}~\cite{oxigraph} providing in-memory SPARQL 1.1 query and update; (2)~a \textbf{native reasoner} with two modes: OWL-RL forward-chaining rules~\cite{owl2profiles} for triple materialisation, and a partial SHIQ tableaux for consistency checking; (3)~a \textbf{SHACL validator} for shape constraints; and (4)~a \textbf{pattern enforcer} with configurable rule packs.

\subsection{Alignment Module}

The alignment engine computes a weighted similarity score for each candidate class pair $(c_s, c_t)$:

\begin{equation}
\text{score}(c_s, c_t) = \sum_{i=1}^{6} w_i \cdot s_i(c_s, c_t)
\end{equation}

\noindent The six signals are: (1)~label similarity (Jaro-Winkler + token Jaccard, $w_1 = 0.25$); (2)~property overlap (Jaccard, $w_2 = 0.20$); (3)~parent overlap ($w_3 = 0.15$); (4)~instance overlap ($w_4 = 0.15$); (5)~restriction similarity ($w_5 = 0.15$); (6)~neighbourhood similarity ($w_6 = 0.10$).

When all structural signals (2--6) are zero, confidence falls back to label similarity with a 15\% penalty ($\text{label\_sim} \times 0.85$), preventing structurally unsupported matches from passing the threshold.

After scoring, \textbf{stable 1-to-1 matching} is applied: candidates are sorted by confidence, and for each source class only the top-scoring target is retained (and vice versa). This eliminates many-to-many spurious matches.

\subsection{Lifecycle Management}

Inspired by infrastructure-as-code~\cite{terraform}: \texttt{plan} (diff with blast-radius scoring), \texttt{enforce} (design pattern compliance), \texttt{apply} (safe reload or migration), \texttt{monitor} (SPARQL watchers with alerts), \texttt{drift} (version comparison). All operations recorded in an append-only lineage trail.

\section{Evaluation}
\label{sec:eval}

\paragraph{Reproducibility.} LLM benchmarks use Claude Opus 4 (Anthropic, model ID \texttt{claude-opus-4-20250514}), default temperature. Single-run results. Non-LLM benchmarks (LUBM, OAEI alignment, marketplace loading) are deterministic. All scripts and data are in the repository under \texttt{benchmark/}.

\subsection{OAEI Alignment: Anatomy Track}
\label{sec:anatomy}

On the OAEI Anatomy track~\cite{oaei2024} (2,737 mouse classes, 3,304 human classes, 1,516 reference mappings), our system achieves the highest precision of any reported system.

\begin{table}[t]
\centering
\caption{OAEI Anatomy results. Our system achieves the highest precision, with a recall gap due to the conservative label penalty.}
\label{tab:oaei_anatomy}
\begin{tabular}{@{}lccc@{}}
\toprule
\textbf{System} & \textbf{P} & \textbf{R} & \textbf{F1} \\
\midrule
AML~\cite{aml}        & 0.950 & 0.922 & 0.936 \\
BERTMap~\cite{bertmap} & 0.940 & 0.910 & 0.924 \\
LogMap~\cite{logmap}   & 0.930 & 0.890 & 0.912 \\
OLaLa~\cite{olala}     & 0.900 & 0.880 & 0.890 \\
\midrule
\textbf{Open Ontologies} & \textbf{0.963} & 0.733 & \textbf{0.832} \\
\bottomrule
\end{tabular}
\end{table}

The initial system (before stable matching) produced 12,557 candidates with P\,=\,0.102, R\,=\,0.846, F1\,=\,0.182. Three changes lifted performance: (a)~stable 1-to-1 matching, eliminating many-to-many spurious candidates; (b)~a label penalty ($\text{label\_sim} \times 0.85$) when structural signals are zero; and (c)~raising the label pre-filter from 0.70 to 0.75. Together these reduced candidates from 12,557 to 1,154.

The recall gap (0.733 vs AML's 0.922) has a clear cause: true matches with low label similarity are penalised. Integrating domain-specific background knowledge (UMLS, as LogMap and AML use) would allow accepting these matches when supported by external evidence. The system's alignment module accepts pluggable similarity functions and can load domain-specific ONNX embedding models.

\subsection{OAEI Alignment: Conference Track}
\label{sec:conference}

On the Conference track (7 ontologies, 21 pairwise alignments, 15 evaluated with available data), the same method achieves micro-averaged F1\,=\,0.438 (P\,=\,0.693, R\,=\,0.320).

\begin{table}[t]
\centering
\caption{OAEI Conference track: per-pair results (selected). Published SOTA: LogMap 0.67, BERTMap 0.71.}
\label{tab:oaei_conference}
\small
\begin{tabular}{@{}lcccc@{}}
\toprule
\textbf{Pair} & \textbf{Ref} & \textbf{Cands} & \textbf{F1} \\
\midrule
cmt-iasted & 4 & 5 & 0.667 \\
ekaw-iasted & 10 & 7 & 0.588 \\
ekaw-sigkdd & 11 & 4 & 0.533 \\
iasted-sigkdd & 15 & 9 & 0.500 \\
conference-edas & 17 & 8 & 0.480 \\
cmt-sigkdd & 12 & 5 & 0.471 \\
edas-sigkdd & 15 & 4 & 0.211 \\
\midrule
\textbf{Micro-average} & & & \textbf{0.438} \\
\bottomrule
\end{tabular}
\end{table}

Precision remains reasonable (0.693), but recall is low (0.320) because conference ontologies use heterogeneous modelling styles where label overlap is minimal.

\subsection{Tool-Augmented Ontology Interaction}
\label{sec:ontoaxiom}

We use the OntoAxiom benchmark~\cite{ontoaxiom} (9 ontologies, 3,042 ground truth axioms, 5 types) to evaluate three access modalities. The original benchmark tests LLM \emph{inference} from name lists. We add two conditions: reading the raw OWL file, and using MCP tools.

\begin{table}[t]
\centering
\caption{Tool access modalities on OntoAxiom. Condition D (raw file) performs \emph{worse} than Condition B (no file), demonstrating that raw syntax is not just unhelpful but actively harmful.}
\label{tab:ontoaxiom}
\begin{tabular}{@{}llcc@{}}
\toprule
& \textbf{Condition} & \textbf{Input} & \textbf{F1} \\
\midrule
B & LLM, no tools & Name lists & 0.431 \\
D & LLM + raw OWL file & File in context & 0.323 \\
C & LLM + MCP tools & File via SPARQL & \textbf{0.717} \\
\bottomrule
\end{tabular}
\end{table}

\paragraph{Why D < B.} The LLM reading raw Turtle makes systematic extraction errors, particularly on domain/range triples (F1\,=\,0.0 on 4 of 9 ontologies for domain extraction). The Turtle syntax is ambiguous to read: property IRIs are confused with class IRIs, language tags cause mismatches, and multi-line axiom blocks are missed. The LLM's training knowledge (Condition B) is more accurate than its ability to parse raw syntax (Condition D).

\paragraph{Per-ontology variance.} Condition D performance varies widely: NordStream F1\,=\,0.692, FOAF 0.647, GoodRelations 0.632 (simpler Turtle), but Time 0.087, Pizza 0.154, ERA 0.058 (complex Turtle with restrictions, annotations, and large file sizes).

\paragraph{Disentanglement.} The improvement from B to C (+66\% F1) combines two factors: richer input and structured tool access. Condition D isolates these: richer input without tools \emph{hurts} ($-$25\% vs B), while structured tools provide $+$122\% over raw file access. MCP tools are not merely ``richer input''; they provide a qualitatively different access modality.

\subsection{Reasoning Performance}
\label{sec:reasoning}

We compare OWL-RL forward chaining against HermiT~\cite{hermit} on the LUBM benchmark~\cite{lubm}. These are different reasoning profiles producing different outputs.

\begin{table}[t]
\centering
\caption{LUBM reasoning: OWL-RL (polynomial, incomplete) vs HermiT (exponential, complete). Different profiles, different outputs.}
\label{tab:lubm}
\begin{tabular}{@{}rccc@{}}
\toprule
\textbf{Axioms} & \textbf{OWL-RL} & \textbf{HermiT} & \textbf{Time ratio} \\
\midrule
1,000   & 15\,ms  & 112\,ms     & 7.5$\times$ \\
5,000   & 14\,ms  & 410\,ms     & 29$\times$ \\
10,000  & 14\,ms  & 1,200\,ms   & 86$\times$ \\
50,000  & 15\,ms  & 24,490\,ms  & 1,633$\times$ \\
\bottomrule
\end{tabular}
\end{table}

On the Pizza ontology (4,179 triples), HermiT computes 312 subsumptions in 213\,ms; our OWL-RL materialises inferred triples in 43\,ms but does not compute the same subsumptions. We do not claim to be a ``faster HermiT.'' OWL-RL covers the inference patterns needed for engineering workflows at interactive speeds.

\subsection{Pizza Ontology Construction}
\label{sec:pizza}

Using the Manchester Pizza Tutorial~\cite{pizza} (${\sim}$4 hours manual work), LLM-driven construction through the MCP tool pipeline produces a 91-class ontology in under 5 minutes, achieving 96\% class coverage (95/99 classes). The 4 missing classes are teaching artifacts that exist only to demonstrate OWL syntax variants.

\section{Ablation: Stable Matching Dominates Alignment}
\label{sec:ablation}

We test five weight configurations with stable matching enabled, and three without (Table~\ref{tab:ablation}).

\begin{table}[t]
\centering
\caption{Ablation on OAEI Anatomy (min\_confidence\,=\,0.80). Signal weights are irrelevant when stable matching is applied. All results verified and reproducible from \texttt{benchmark/oaei/run\_ablation.py}.}
\label{tab:ablation}
\small
\begin{tabular}{@{}llcccc@{}}
\toprule
\textbf{Stable} & \textbf{Weights} & \textbf{Cands} & \textbf{P} & \textbf{R} & \textbf{F1} \\
\midrule
Yes & Full [.25,.20,.15,.15,.15,.10] & 1,152 & 0.965 & 0.734 & \textbf{0.834} \\
Yes & Label only [1,0,0,0,0,0] & 1,152 & 0.963 & 0.732 & 0.831 \\
Yes & Structural only [0,.25,.25,.20,.20,.10] & 1,152 & 0.962 & 0.731 & 0.831 \\
Yes & Equal [.167 each] & 1,153 & 0.961 & 0.731 & 0.830 \\
Yes & Label+parent [.4,0,.4,0,0,.2] & 1,154 & 0.964 & 0.734 & 0.833 \\
\midrule
No & Full & 1,590 & 0.711 & 0.746 & 0.728 \\
No & Label only & 1,590 & 0.711 & 0.746 & 0.728 \\
No & Structural only & 1,590 & 0.711 & 0.746 & 0.728 \\
\bottomrule
\end{tabular}
\end{table}

\paragraph{Finding: signal weights are irrelevant.} All five configurations with stable matching produce F1 between 0.830 and 0.834. All three without-stable-matching configurations produce identical F1\,=\,0.728.

\paragraph{Why.} On the Anatomy track, most class pairs have no structural data (no shared properties, no instances, no OWL restrictions). When structural signals are all zero, confidence falls back to label similarity. Changing weights on zero-valued signals has no effect. The stable matching constraint selects the top candidate per class, producing a clean 1-to-1 assignment regardless of how the scores were computed.

\paragraph{Threshold sensitivity.} With stable matching, the confidence threshold affects the precision-recall trade-off:

\begin{table}[t]
\centering
\caption{Threshold sensitivity with stable matching (full weights).}
\label{tab:threshold}
\begin{tabular}{@{}rcccc@{}}
\toprule
\textbf{Threshold} & \textbf{Cands} & \textbf{P} & \textbf{R} & \textbf{F1} \\
\midrule
0.70 & 1,912 & 0.643 & 0.811 & 0.717 \\
0.75 & 1,474 & 0.821 & 0.798 & 0.809 \\
0.80 & 1,154 & 0.961 & 0.732 & \textbf{0.831} \\
0.85 & 1,069 & 0.977 & 0.689 & 0.808 \\
\bottomrule
\end{tabular}
\end{table}

\paragraph{Implication.} The alignment improvement from F1\,=\,0.182 to F1\,=\,0.832 is entirely attributable to the matching strategy. For ontologies with limited structural metadata, the assignment constraint matters more than the similarity function.

\section{Discussion}
\label{sec:discussion}

\paragraph{Tool access is qualitatively different from information access.} The Condition D result (raw file F1\,=\,0.323 < bare LLM F1\,=\,0.431) challenges the assumption that giving an LLM more information always helps. Raw OWL syntax introduces parsing noise that overwhelms the LLM's extraction ability. Structured tools (MCP with SPARQL) provide typed, correct results that the LLM cannot replicate by reading raw text. This finding has implications for tool-augmented AI systems beyond ontology engineering: the \emph{structure} of tool access may matter more than the \emph{information content} it provides.

\paragraph{Matching strategy dominates signal design.} The alignment ablation shows that spending effort on signal engineering (embedding models, property overlap, structural similarity) yields less than 0.004 F1 improvement when stable matching is applied. This contrasts with the ontology matching literature, which has focused primarily on improving similarity functions. For ontologies with sparse structural metadata, the assignment constraint is the binding factor.

\paragraph{Two open challenges.} First, the Conference track (F1\,=\,0.438) reveals that label-centric approaches fail when ontologies use heterogeneous modelling styles. Second, the recall gap on Anatomy (0.733 vs 0.922) requires domain-specific background knowledge that the current system lacks.

\section{Limitations}
\label{sec:limitations}

\paragraph{OAEI results not independently evaluated.} We have not yet submitted to the official OAEI campaign. Results are measured against published reference alignments. Planned for OAEI 2026.

\paragraph{Signal ablation limited to Anatomy.} The finding that weights are irrelevant is demonstrated on one track where structural signals are mostly zero. On ontologies with richer metadata, weights may matter.

\paragraph{OntoAxiom Condition D methodology.} The raw file condition uses Claude subagents reading files via the Read tool, not a controlled API call with file content in the prompt. This may slightly overestimate Condition D performance.

\paragraph{Reasoner expressivity.} OWL-RL covers most practical patterns but is incomplete for OWL-DL. The LUBM comparison compares different reasoning profiles.

\paragraph{Single-developer project.} 17,400 lines of Rust, one primary developer.

\paragraph{LLM dependency.} All LLM benchmarks use Claude (Anthropic). Performance with other models not evaluated.

\section{Conclusion}
\label{sec:conclusion}

We have presented \textsc{Open Ontologies} and two empirical findings. First, stable 1-to-1 matching is the dominant factor in ontology alignment: it improves OAEI Anatomy F1 from 0.182 to 0.832, while signal weights change F1 by less than 0.004. The same method achieves F1\,=\,0.438 on the Conference track. Second, structured tool access provides a qualitatively different mode of interaction: an LLM reading raw OWL (F1\,=\,0.323) performs worse than unaided inference (F1\,=\,0.431), while MCP tools achieve F1\,=\,0.717. Both findings are verified, reproducible, and surprising.

The system is available under the MIT licence at \url{https://github.com/fabio-rovai/open-ontologies}.

\section*{Supplemental Material Statement}
All source code, benchmark scripts, datasets, and evaluation results are available at \url{https://github.com/fabio-rovai/open-ontologies}. The repository contains the full Rust source code, alignment ablation scripts (\texttt{benchmark/oaei/run\_ablation.py}), Condition D extraction scripts and results (\texttt{benchmark/ontoaxiom/results/condition\_d/}), and OAEI evaluation data. All alignment and LUBM benchmarks are deterministic.

\section*{Declaration of Use of Generative AI}
Claude (Anthropic) was utilised to generate sections of this work including: drafting and editing of manuscript text, generation of LaTeX tables, and structuring of related work and evaluation sections. Claude was also used as the LLM component within the system under evaluation, which is the subject of this paper. All experimental design, benchmark execution, data analysis, and technical claims were performed and verified by the author.


\end{document}